\documentclass[11pt,letterpaper]{article}
\usepackage{naaclhlt2015}
\usepackage{times}
\usepackage{latexsym}
\usepackage{amsmath}
\usepackage{amssymb}
\usepackage{hyperref}
\usepackage{graphicx}
\usepackage{multirow}
\usepackage[round]{natbib}
\usepackage{url}
\usepackage{booktabs}

\hypersetup{
    colorlinks=false,
    pdfborder={0 0 0},
}

\title{Pragmatic Neural Language Modelling in Machine Translation}

\author{Paul Baltescu \\
	    University of Oxford\\
	    {\tt paul.baltescu@cs.ox.ac.uk }
	  \And
	Phil Blunsom\\
  	University of Oxford\\
  	Google DeepMind\\
  {\tt phil.blunsom@cs.ox.ac.uk }}

\date{}

\begin{document}
\maketitle

\begin{abstract}



This paper presents an in-depth investigation on integrating neural language models in translation systems. Scaling neural language models is a difficult task, but crucial for real-world applications. This paper evaluates the impact on end-to-end MT quality of both new and existing scaling techniques. We show when explicitly normalising neural models is necessary and what optimisation tricks one should use in such scenarios. We also focus on scalable training algorithms and investigate noise contrastive estimation and diagonal contexts as sources for further speed improvements. We explore the trade-offs between neural models and back-off n-gram models and find that neural models make strong candidates for natural language applications in memory constrained environments, yet still lag behind traditional models in raw translation quality. We conclude with a set of recommendations one should follow to build a scalable neural language model for MT.

\end{abstract}
\section{Introduction}

Language models are used in translation systems to improve the fluency of the output translations. The most popular language model implementation is a back-off n-gram model with Kneser-Ney smoothing \citep{Chen1999}. Back-off n-gram models are conceptually simple, very efficient to construct and query, and are regarded as being extremely effective in translation systems.

Neural language models are a more recent class of language models \citep{Bengio2003} that have been shown to outperform back-off n-gram models using intrinsic evaluations of heldout perplexity \citep{Chelba2013, Bengio2003}, or when used in addition to traditional models in natural language systems such as speech recognizers \citep{Mikolov2011, Schwenk2007}. Neural language models combat the problem of data sparsity inherent to traditional n-gram models by learning distributed representations for words in a continuous vector space. 



It has been shown that neural language models can improve translation quality when used as additional features in a decoder \citep{Vaswani2013, Botha2014, Baltescu2014, Auli2014} or if used for n-best list rescoring \citep{Schwenk2010, Auli2013}. These results show great promise and in this paper we continue this line of research by investigating the trade-off between speed and accuracy when integrating neural language models in a decoder. We also focus on how effective these models are when used as the sole language model in a translation system. This is important because our hypothesis is that most of the language modelling is done by the n-gram model, with the neural model only acting as a differentiating factor when the n-gram model cannot provide a decisive probability. Furthermore, neural language models are considerably more compact and represent strong candidates for modelling language in memory constrained environments (e.g. mobile devices, commodity machines, etc.), where back-off n-gram models trained on large amounts of data do not fit into memory. 

Our results show that a novel combination of noise contrastive estimation \citep{Mnih2012} and factoring the softmax layer using Brown clusters \citep{Brown1992} provides the most pragmatic solution for fast training and decoding. Further, we confirm that when evaluated purely on BLEU score, neural models are unable to match the benchmark Kneser-Ney models, even if trained with large hidden layers. However, when the evaluation is restricted to models that match a certain memory footprint, neural models clearly outperform the n-gram benchmarks, confirming that they represent a practical solution for memory constrained environments.


\section{Model Description}

\begin{figure}
\includegraphics[scale=0.28]{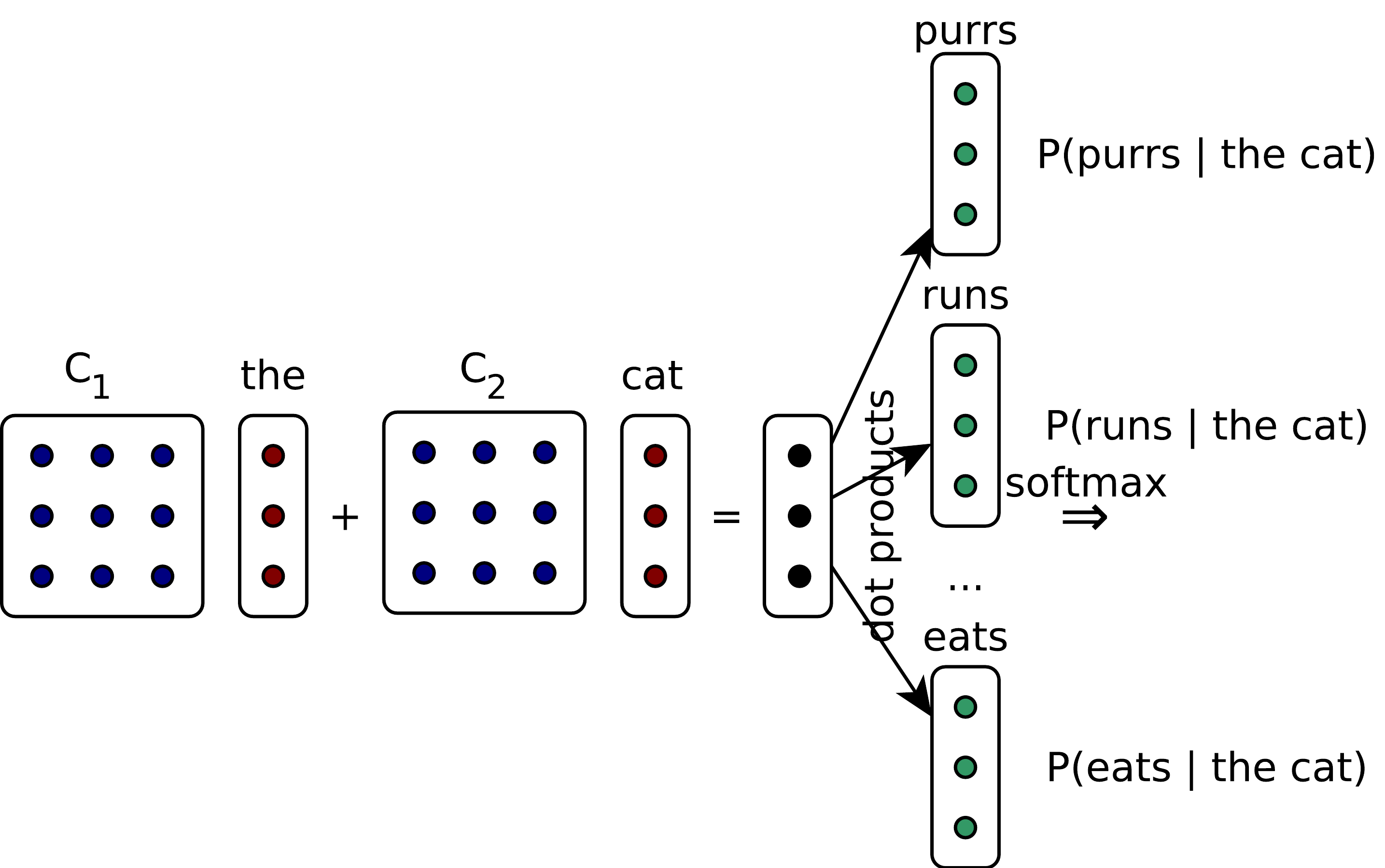}
\caption{A 3-gram neural language model is used to predict the word following the context \textit{the cat}.}
\label{fig:architecture}
\end{figure}

As a basis for our investigation, we implement a probabilistic neural language model as defined in \cite{Bengio2003}.\footnote{Our goal is to release a scalable neural language modelling toolkit at the following URL: \url{http://www.example.com}.} For every word $w$ in the vocabulary $V$, we learn two distributed representations $\textbf{q}_w$ and $\textbf{r}_w$ in $\mathbb{R}^D$. The vector $\textbf{q}_w$ captures the syntactic and semantic role of the word $w$ when $w$ is part of a conditioning context, while $\textbf{r}_w$ captures its role as a prediction. For some word $w_i$ in a given corpus, let $h_i$ denote the conditioning context $w_{i-1}, \ldots, w_{i-n+1}$. To find the conditional probability $P(w_i | h_i)$, our model first computes a context projection vector:
\begin{equation*}
\textbf{p} = f\left( \sum_{j=1}^{n-1} \text{C}_j \textbf{q}_{h_{ij}} \right),
\end{equation*}
where $\text{C}_j \in \mathbb{R}^{D \times D}$ are context specific transformation matrices and $f$ is a component-wise \textit{rectified linear} activation. The model computes a set of similarity scores measuring how well each word $w \in V$ matches the context projection of $h_i$. The similarity score is defined as $\phi(w, h_i) = \textbf{r}_w^\text{T} \textbf{p} + b_w$, where $b_w$ is a bias term incorporating the prior probability of the word $w$. The similarity scores are transformed into probabilities using the softmax function:
\begin{equation*}
P(w_i | h_i) = \frac{\exp (\phi(w_i, h_i))}{\sum_{w \in V} \exp (\phi(w, h_i))},
\end{equation*}
The model architecture is illustrated in \autoref{fig:architecture}. The parameters are learned with gradient descent to maximize log-likelihood with $L_2$ regularization.

\begin{table}[t]
\begin{center}
\begin{tabular}{ccc}
\toprule
\textbf{Model} & \textbf{Training} & \textbf{Exact Decoding} \\
\midrule
Standard & \small{$O(|V|\times D)$} & \small{$O(|V|\times D)$} \\
Class Factored & \small{$O(\sqrt{|V|}\times D)$} & \small{$O(\sqrt{|V|}\times D)$} \\
Tree Factored & \small{$O(\log|V|\times D)$} & \small{$O(\log{|V|} \times D)$} \\
NCE & \small{$O(k\times D)$} & \small{$O(|V| \times D)$} \\
\bottomrule
\end{tabular}
\end{center}
\vspace{-0.3cm}
\caption{Training and decoding complexities for the optimization tricks discussed in section 2.}
\label{table:complexities}
\end{table}

Scaling neural language models is hard because any forward pass through the underlying neural network computes an expensive softmax activation in the output layer. This operation is performed during training and testing for all contexts presented as input to the network. Several methods have been proposed to alleviate this problem: some applicable only during training \citep{Mnih2012, Bengio2008}, while others may also speed up arbitrary queries to the language model \citep{Morin2005, Mnih2009}.

In the following subsections, we present several extensions to this model, all sharing the goal of reducing the computational cost of the softmax step. \autoref{table:complexities} summarizes the complexities of these methods during training and decoding.

\subsection{Class Based Factorisation}

The time complexity of the softmax step is $O(|V| \times D)$. One option for reducing this excessive amount of computation is to rely on a class based factorisation trick \citep{Goodman2001}. We partition the vocabulary into $K$ classes $\{\mathcal{C}_1, \ldots, \mathcal{C}_K\}$ such that $V = \bigcup_{i=1}^K \mathcal{C}_i$ and $\mathcal{C}_i \cap \mathcal{C}_j = \varnothing, \forall 1 \leq i < j \leq K$. We define the conditional probabilities as:
\begin{equation*}
P(w_i | h_i) = P(c_i | h_i) P(w_i | c_i, h_i),
\end{equation*}
where $c_i$ is the class the word $w_i$ belongs to, i.e. $w_i \in \mathcal{C}_{c_i}$. We adjust the model definition to also account for the class probabilities $P(c_i | h_i)$. We associate a distributed representation $\textbf{s}_c$ and a bias term $t_c$ to every class $c$. The class conditional probabilities are computed reusing the projection vector $\textbf{p}$ with a new scoring function $\psi(c, h_i) = \textbf{s}_c^\text{T} \textbf{p} + t_c$. The probabilities are normalised separately:
\begin{align*}
P(c_i | h_i) &= \frac{\exp(\psi(c_i, h_i))}{\sum_{j=1}^K \exp(\psi(c_j, h_i))} \\
P(w_i | c_i, h_i) &= \frac{\exp(\phi(w_i, h_i))}{\sum_{w \in \mathcal{C}_{c_i}} \exp(\phi(w, h_i))}
\end{align*}
When $K \approx \sqrt{|V|}$ and the word classes have roughly equal sizes, the softmax step has a more manageable time complexity of $O(\sqrt{|V|} \times D)$ for both training and testing.

\subsection{Tree Factored Models}

One can take the idea presented in the previous section one step further and construct a tree over the vocabulary $V$. The words in the vocabulary are used to label the leaves of the tree. Let $n_1, \ldots, n_k$ be the nodes on the path descending from the root ($n_1$) to the leaf labelled with $w_i$ ($n_k$). The probability of the word $w_i$ to follow the context $h_i$ is defined as:
\begin{equation*}
P(w_i | h_i) = \prod_{j=2}^k P(n_j | n_1, \ldots, n_{j-1}, h_i).
\end{equation*}
We associate a distributed representation $\textbf{s}_n$ and bias term $t_n$ to each node in the tree. The conditional probabilities are obtained reusing the scoring function $\psi(n_j, h_i)$:
\begin{equation*}
P(n_j | n_1, \ldots, n_{j-1}, h_i) = \frac{\exp(\psi(n_j, h_i))}{\sum_{n \in \mathcal{S}(n_j)} \exp(\psi(n, h_i))},
\end{equation*}
where $\mathcal{S}(n_j)$ is the set containing the siblings of $n_j$ and the node itself. Note that the class decomposition trick described earlier can be understood as a tree factored model with two layers, where the first layer contains the word classes and the second layer contains the words in the vocabulary.

The optimal time complexity is obtained by using balanced binary trees. The overall complexity of the normalisation step becomes $O(\log |V| \times D)$ because the length of any path is bounded by $O(\log |V|)$ and because exactly two terms are present in the denominator of every normalisation operation. 

Inducing high quality binary trees is a difficult problem which has received some attention in the research literature \citep{Mnih2009, Morin2005}. Results have been somewhat unsatisfactory, with the exception of \cite{Mnih2009}, who did not release the code they used to construct their trees. In our experiments, we use Huffman trees \citep{Huffman1952} which do not have any linguistic motivation, but guarantee that a minimum number of nodes are accessed during training. Huffman trees have depths that are close to $\log{|V|}$.

\subsection{Noise Contrastive Estimation}

Training neural language models to maximise data likelihood involves several iterations over the entire training corpus and applying the backpropagation algorithm for every training sample. Even with the previous factorisation tricks, training neural models is slow. We investigate an alternative approach for training language models based on noise contrastive estimation, a technique which does not require normalised probabilities when computing gradients \citep{Mnih2012}. This method has already been used for training neural language models for machine translation by \cite{Vaswani2013}.

The idea behind noise contrastive training is to transform a density estimation problem into a classification problem, by learning a classifier to discriminate between samples drawn from the data distribution and samples drawn for a known noise distribution. Following \cite{Mnih2012}, we set the unigram distribution $P_n(w)$ as the noise distribution and use $k$ times more noise samples than data samples to train our models. The new objective is:
\begin{align*}
J(\theta) &= \sum_{i=1}^m \log P(C = 1 | \theta, w_i, h_i) \\
          &+ \sum_{i=1}^m \sum_{j=1}^{k} \log P(C = 0 | \theta, n_{ij}, h_i),
\end{align*}
where $n_{ij}$ are the noise samples drawn from $P_n(w)$. The posterior probability that a word is generated from the data distribution given its context is:
\begin{equation*}
P(C = 1 | \theta, w_i, h_i) = \frac{P(w_i | \theta, h_i)}{P(w_i | \theta, h_i) + k P_n(w_i)}.
\end{equation*}
\cite{Mnih2012} show that the gradient of $J(\theta)$ converges to the gradient of the log-likelihood objective when $k \rightarrow \infty$. 

When using noise contrastive estimation, additional parameters can be used to capture the normalisation terms. \cite{Mnih2012} fix these parameters to 1 and obtain the same perplexities, thereby circumventing the need for explicit normalisation. However, this method does not provide any guarantees that the models are normalised at test time. In fact, the outputs may sum up to arbitrary values, unless the model is explicitly normalised.

Noise contrastive estimation is more efficient than the factorisation tricks at training time, but at test time one still has to normalise the model to obtain valid probabilities. We propose combining this approach with the class decomposition trick resulting in a fast algorithm for both training and testing. In the new training algorithm, when we account for the class conditional probabilities $P(c_i | h_i)$, we draw noise samples from the class unigram distribution, and when we account for $P(w_i | c_i, h_i)$, we sample from the unigram distribution of only the words in the class $\mathcal{C}_{c_i}$.

\section{Experimental Setup}

In our experiments, we use data from the 2014 ACL Workshop in Machine Translation.\footnote{The data is available here: \url{http://www.statmt.org/wmt14/translation-task.html}.} We train standard phrase-based translation systems for French $\rightarrow$ English, English $\rightarrow$ Czech and English $\rightarrow$ German using the \texttt{Moses} toolkit \citep{Koehn2007}.

\begin{table}
\begin{center}
\begin{tabular}{ccc}
\toprule
\textbf{Language pairs} & \textbf{\# tokens} & \textbf{\# sentences} \\
\midrule
fr$\rightarrow$en & 113M & 2M \\
en$\rightarrow$cs & 36.5M & 733.4k \\
en$\rightarrow$de & 104.9M & 1.95M \\
\bottomrule
\end{tabular}
\end{center}
\caption{Statistics for the parallel corpora.}
\label{table:parallel_data}
\end{table}

\begin{table}
\begin{center}
\begin{tabular}{ccc}
\toprule
\textbf{Language} & \textbf{\# tokens} & \textbf{Vocabulary} \\
\midrule
English (en) & 2.05B & 105.5k \\
Czech (cs) & 566M & 214.9k \\
German (de) & 1.57B & 369k \\
\bottomrule
\end{tabular}
\end{center}
\caption{Statistics for the monolingual corpora.}
\label{table:monolingual_data}
\end{table}

We used the \texttt{europarl} and the \texttt{news commentary} corpora as parallel data for training the translation systems. The parallel corpora were tokenized, lowercased and sentences longer than 80 words were removed using standard text processing tools.\footnote{We followed the first two steps from \url{http://www.cdec-decoder.org/guide/tutorial.html}.} \autoref{table:parallel_data} contains statistics about the training corpora after the preprocessing step. We tuned the translation systems on the \texttt{newstest2013} data using minimum error rate training \citep{Och2003b} and we used the \texttt{newstest2014} corpora to report uncased BLEU scores averaged over 3 runs.

The monolingual training data used for training language models consists of the \texttt{europarl}, \texttt{news commentary} and the \texttt{news crawl 2007-2013} corpora. The corpora were tokenized and lowercased using the same text processing scripts and the words not occuring the in the target side of the parallel data were replaced with a special \texttt{<unk>} token. Statistics for the monolingual data after the preprocessing step are reported in \autoref{table:monolingual_data}. 

Throughout this paper we report results for 5-gram language models, regardless of whether they are back-off n-gram models or neural models. To construct the back-off n-gram models, we used a compact trie-based implementation available in \texttt{KenLM} \citep{Heafield2011}, because otherwise we would have had difficulties with fitting these models in the main memory of our machines. When training neural language models, we set the size of the distributed representations to 500, we used diagonal context matrices and we used 10 negative samples for noise contrastive estimation, unless otherwise indicated. In cases where we perform experiments on only one language pair, the reader should assume we used French$\rightarrow$English data.
\section{Normalisation}

\begin{table}[t]
\begin{center}
\begin{tabular}{cccc}
\toprule
\textbf{Model} & \textbf{fr$\rightarrow$en} & \textbf{en$\rightarrow$cs} & \textbf{en$\rightarrow$de} \\
\midrule
KenLM & 33.01 \footnotesize{(120.446)} & 19.11 & 19.75 \\
NLM  & 31.55 \footnotesize{(115.119)} & 18.56 & 18.33 \\
\bottomrule
\end{tabular}
\end{center}
\vspace{-0.3cm}
\caption{A comparison between standard back-off n-gram models and neural language models. The perplexities for the English language models are shown in parentheses.}
\label{table:comparison}
\end{table}

The key challenge with neural language models is scaling the softmax step in the output layer of the network. This operation is especially problematic when the neural language model is incorporated as a feature in the decoder, as the language model is queried several hundred thousand times for any sentence of average length.

Previous publications on neural language models in machine translation have approached this problem in two different ways. \cite{Vaswani2013} and \cite{Devlin2014} simply ignore normalisation when decoding, albeit \cite{Devlin2014} alter their training objective to learn self-normalised models, i.e. models where the sum of the values in the output layer is (hopefully) close to 1. \cite{Vaswani2013} use noise contrastive estimation to speed up training, while \cite{Devlin2014} train their models with standard gradient descent on a GPU.

The second approach is to explicitly normalise the models, but to limit the set of words over which the normalisation is performed, either via class-based factorisation \citep{Botha2014, Baltescu2014} or using a shortlist containing only the most frequent words in the vocabulary and scoring the remaining words with a back-off n-gram model \citep{Schwenk2010}. Tree factored models follow the same general approach, but to our knowledge, they have never been investigated in a translation system before. These normalisation techniques can be successfully applied both when training the models and when using them in a decoder.

\autoref{table:comparison} shows a side by side comparison of out of the box neural language models and back-off n-gram models. We note a significant drop in quality when neural language models are used (roughly 1.5 BLEU for fr$\rightarrow$en and en$\rightarrow$de and 0.5 BLEU for en$\rightarrow$ cs). This result is in line with \cite{Zhao2014} and shows that by default back-off n-gram models are much more effective in MT. An interesting observation is that the neural models have lower perplexities than the n-gram models, implying that BLEU scores and perplexities are only loosely correlated.

\begin{table}[t]
\begin{center}
\begin{tabular}{cccc}
\toprule
\textbf{Normalisation} & \textbf{fr$\rightarrow$en} & \textbf{en$\rightarrow$cs} & \textbf{en$\rightarrow$de} \\
\midrule
Unnormalised & 33.89 & 20.06 & 20.25 \\
Class Factored & 33.87 & 19.96 & 20.25 \\
Tree Factored & 33.69 & 19.52 & 19.87 \\
\bottomrule
\end{tabular}
\end{center}
\vspace{-0.3cm}
\caption{Qualitative analysis of the proposed normalisation schemes \textit{with} an additional back-off n-gram model.}
\label{table:withkenlm}
\end{table}

\begin{table}[t]
\begin{center}
\begin{tabular}{cccc}
\toprule
\textbf{Normalisation} & \textbf{fr$\rightarrow$en} & \textbf{en$\rightarrow$cs} & \textbf{en$\rightarrow$de} \\
\midrule
Unnormalised & 30.98 & 18.57 & 18.05 \\
Class Factored  & 31.55 & 18.56 & 18.33 \\
Tree Factored & 30.37 & 17.19 & 17.26 \\
\bottomrule
\end{tabular}
\end{center}
\vspace{-0.3cm}
\caption{Qualitative analysis of the proposed normalisation schemes \textit{without} an additional back-off n-gram model.}
\label{table:nokenlm}
\end{table}

\autoref{table:withkenlm} and \autoref{table:nokenlm} show the impact on translation quality for the proposed normalisation schemes with and without an additional n-gram model. We note that when \texttt{KenLM} is used, no significant differences are observed between normalised and unnormalised models, which is again in accordance with the results of \cite{Zhao2014}. However, when the n-gram model is removed, class factored models perform better (at least for fr$\rightarrow$en and en$\rightarrow$de), despite being only an approximation of the fully normalised models. We believe this difference in not observed in the first case because most of the language modelling is done by the n-gram model (as indicated by the results in \autoref{table:comparison}) and that the neural models only act as a differentiating feature when the n-gram models do not provide accurate probabilities. We conclude that some form of normalisation is likely to be necessary whenever neural models are used alone. This result may also explain why \cite{Zhao2014} show, perhaps surprisingly, that normalisation is important when reranking n-best lists with recurrent neural language models, but not in other cases. (This is the only scenario where they use neural models without supporting n-gram models.)

\autoref{table:withkenlm} and \autoref{table:nokenlm} also show that tree factored models perform poorly compared to the other candidates. We believe this is likely to be a result of the artificial hierarchy imposed by the tree over the vocabulary.

\begin{table}[t]
\begin{center}
\begin{tabular}{ccc}
\toprule
\textbf{Normalisation} & \textbf{Clustering} & \textbf{BLEU} \\
\midrule
Class Factored & Brown clustering & 31.55 \\
Class Factored & Frequency binning & 31.07 \\
Tree Factored & Huffman encoding & 30.37 \\
\bottomrule
\end{tabular}
\end{center}
\vspace{-0.3cm}
\caption{Qualitative analysis of clustering strategies on fr$\rightarrow$en data.}
\label{table:clustering}
\end{table}

\begin{table}[t]
\begin{center}
\begin{tabular}{cc}
\toprule
\textbf{Model} & \textbf{Average decoding time} \\
\midrule
KenLM & 1.64 s \\
Unnormalised NLM & 3.31 s \\
Class Factored NLM & 42.22 s \\
Tree Factored NLM & 18.82 s \\
\bottomrule
\end{tabular}
\end{center}
\vspace{-0.3cm}
\caption{Average decoding time per sentence for the proposed normalisation schemes.}
\label{table:decoding}
\end{table}

\autoref{table:clustering} compares two popular techniques for obtaining word classes: Brown clustering \citep{Brown1992, Liang2005} and frequency binning \citep{Mikolov2011c}. From these results, we learn that the clustering technique employed to partition the vocabulary into classes can have a huge impact on translation quality and that Brown clustering is clearly superior to frequency binning. 

Another thing to note is that frequency binning partitions the vocabulary in a similar way to Huffman encoding. This observation implies that the BLEU scores we report for tree factored models are not optimal, but we can get an insight on how much we expect to lose in general by imposing a tree structure over the vocabulary (on the fr$\rightarrow$en setup, we lose roughly 0.7 BLEU points). Unfortunately, we are not able to report BLEU scores for factored models using Brown trees because the time complexity for constructing such trees is $O(|V|^3)$.

We report the average time needed to decode a sentence for each of the models described in this paper in  \autoref{table:decoding}. We note that factored models are slow compared to unnormalised models. One option for speeding up factored models is using a GPU to perform the vector-matrix operations. However, GPU integration is architecture specific and thus against our goal of making our language modelling toolkit usable by everyone.

\section{Training}

\begin{table}[t]
\begin{center}
\begin{tabular}{cccc}
\toprule
\textbf{Training} & \textbf{Perplexity} & \textbf{BLEU} & \textbf{Duration} \\
\midrule
SGD & 116.596 & 31.75 & 9.1 days \\
NCE & 115.119 & 31.55 & 1.2 days \\
\bottomrule
\end{tabular}
\end{center}
\vspace{-0.3cm}
\caption{A comparison between stochastic gradient descent (SGD) and noise contrastive estimation (NCE) for class factored models on the fr$\rightarrow$en data.}
\label{table:nce}
\end{table}

\begin{table}[t]
\begin{center}
\begin{tabular}{cc}
\toprule
\textbf{Model} & \textbf{Training time} \\
\midrule
Unnormalised NCE & 1.23 days \\
Class Factored NCE & 1.20 days \\
Tree Factored SGD & 1.4 days \\
\bottomrule
\end{tabular}
\end{center}
\vspace{-0.3cm}
\caption{Training times for neural models on fr$\rightarrow$en data.}
\label{table:training}
\end{table}

In this section, we are concerned with finding scalable training algorithms for neural language models. We investigate noise contrastive estimation as a much more efficient alternative to standard maximum likelihood training via stochastic gradient descent. Class factored models enable us to conduct this investigation at a much larger scale than previous results (e.g. the WSJ corpus used by \cite{Mnih2012} has slightly over 1M tokens), thereby gaining useful insights on how this method truly performs at scale. (In our experiments, we use a 2B words corpus and a 100k vocabulary.) \autoref{table:nce} summarizes our findings. We obtain a slightly better BLEU score with stochastic gradient descent, but this is likely to be just noise from tuning the translation system with MERT. On the other hand, noise contrastive training reduces training time by a factor of 7.

\autoref{table:training} reviews the neural models described in this paper and shows the time needed to train each one. We note that noise contrastive training requires roughly the same amount of time regardless of the structure of the model. Also, we note that this method is at least as fast as maximum likelihood training even when the latter is applied to tree factored models. Since tree factored models have lower quality, take longer to query and do not yield any substantial benefits at training time when compared to unnormalised models, we conclude they represent a suboptimal language modelling choice for machine translation.

\section{Diagonal Context Matrices}

\begin{table}[t]
\begin{center}
\begin{tabular}{cccc}
\toprule
\textbf{Contexts} & \textbf{Perplexity} & \textbf{BLEU} & \textbf{Training time} \\
\midrule
Full & 114.113 & 31.43 & 3.64 days \\
Diagonal & 115.119 & 31.55 & 1.20 days \\
\bottomrule
\end{tabular}
\end{center}
\vspace{-0.3cm}
\caption{A side by side comparison of class factored models with and without diagonal contexts trained with noise contrastive estimation on the fr$\rightarrow$en data.}
\label{table:contexts}
\end{table}

In this section, we investigate diagonal context matrices as a source for reducing the computational cost of calculating the projection vector. In the standard definition of a neural language model, this cost is dominated by the \textit{softmax} step, but as soon as tricks like noise contrastive estimation or tree or class factorisations are used, this operation becomes the main bottleneck for training and querying the model. Using diagonal context matrices when computing the projection layer reduces the time complexity from $O(D^2)$ to $O(D)$.  A similar optimization is achieved in the backpropagation algorithm, as only $O(D)$ context parameters need to be updated for every training instance.

\cite{Devlin2014} also identified the need for finding a scalable solution for computing the projection vector. Their approach is to cache the product between every word embedding and every context matrix and to look up these terms in a table as needed. \cite{Devlin2014}'s approach works well when decoding, but it requires additional memory and is not applicable during training. 

\autoref{table:contexts} compares diagonal and full context matrices for class factored models. Both models have similar BLEU scores, but the training time is reduced by a factor of 3 when diagonal context matrices are used. We obtain similar improvements when decoding with class factored models, but the speed up for unnormalised models is over 100x!

\section{Quality vs. Memory Trade-off}

\begin{figure}
\includegraphics[scale=0.4]{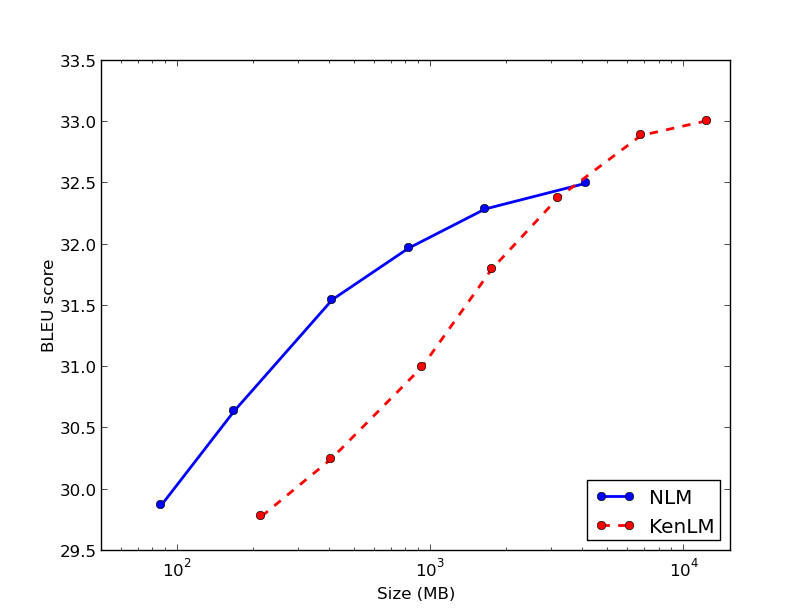}
\caption{A graph highlighting the quality vs. memory trade-off between traditional n-gram models and neural language models.}
\label{fig:memory}
\end{figure}

Neural language models are a very appealing option for natural language applications that are expected to run on mobile phones and commodity computers, where the typical amount of memory available is limited to 1-2 GB. Nowadays, it is becoming more and more common for these devices to include reasonably powerful GPUs, supporting the idea that further scaling is possible if necessary. On the other hand, fitting back-off n-gram models on such devices is difficult because these models store the probability of every n-gram in the training data. In this section, we seek to gain further understanding on how these models perform under such conditions.

In this analysis, we used \cite{Heafield2011}'s trie-based implementation with quantization for constructing memory efficient back-off n-gram models. A 5-gram model trained on the English monolingual data introduced in section 3 requires 12 GB of memory. We randomly sampled sentences with an acceptance ratio ranging between 0.01 and 1 to construct smaller models and observe their performance on a larger spectrum. The BLEU scores obtained using these models are reported in \autoref{fig:memory}. We note that the translation quality improves as the amount of training data increases, but the improvements are less significant when most of the data is used.

The neural language models we used to report results throughout this paper are roughly 400 MB in size. Note that we do not use any compression techniques to obtain smaller models, although this is technically possible (e.g. quantization). We are interested to see how these models perform for various memory thresholds and we experiment with setting the size of the word embeddings between 100 and 5000. More importantly, these experiments are meant to give us an insight on whether very large neural language models have any chance of achieving the same performance as back-off n-gram models in translation tasks. A positive result would imply that significant gains can be obtained by scaling these models further, while a negative result signals a possible inherent inefficiency of neural language models in MT. The results are shown in \autoref{fig:memory}.

From \autoref{fig:memory}, we learn that neural models perform significantly better (over 1 BLEU point) when there is under 1 GB of memory available. This is exactly the amount of memory generally available on mobile phones and ordinary computers, confirming the potential of neural language models for applications designed to run on such devices. However, at the other end of the scale, we can see that back-off models outperform even the largest neural language models by a decent margin and we can expect only modest gains if we scale these models further.

\section{Conclusion}

This paper presents an empirical analysis of neural language models in machine translation. The experiments presented in this paper help us draw several useful conclusions about the ideal usage of these language models in MT systems.

The first problem we investigate is whether normalisation has any impact on translation quality and we survey the effects of some of the most frequently used techniques for scaling neural language models. We conclude that normalisation is not necessary when neural models are used in addition to back-off n-gram models. This result is due to the fact that most of the language modelling is done by the n-gram model. (Experiments show that out of the box n-gram models clearly outperform their neural counterparts.) The MT system learns a smaller weight for neural models and we believe their main use is to correct the inaccuracies of the n-gram models.

On the other hand, when neural language models are used in isolation, we observe that normalisation does matter. We believe this result generalizes to other neural architectures such as neural translation models \citep{Sutskever2014, Cho2014}. We observe that the most effective normalisation strategy in terms of translation quality is the class-based decomposition trick. We learn that the algorithm used for partitioning the vocabulary into classes has a strong impact on the overall quality and that Brown clustering \citep{Brown1992} is a good choice. Decoding with class factored models can be slow, but this issue can be corrected using GPUs, or if a comprise in quality is acceptable, unnormalised models represent a much faster alternative. We also conclude that tree factored models are not a strong candidate for translation since they are outperformed by unnormalised models in every aspect.

We introduce noise contrastive estimation for class factored models and show that it performs almost as well as maximum likelihood training with stochastic gradient descent. To our knowledge, this is the first side by side comparison of these two techniques on a dataset consisting of a few billions of training examples and a vocabulary with over 100k tokens. On this setup, noise contrastive estimation can be used to train standard or class factored models in a little over 1 day.

We explore diagonal context matrices as an optimization for computing the projection layer in the neural network. The trick effectively reduces the time complexity of this operation from $O(D^2)$ to $O(D)$. Compared to \cite{Devlin2014}'s approach of caching vector-matrix products, diagonal context matrices are also useful for speeding up training and do not require additional memory. Our experiments show that diagonal context matrices perform just as well as full matrices in terms of translation quality.

We also explore the trade-off between neural language models and back-off n-gram models. We observe that in the memory range that is typically available on a mobile phone or a commodity computer, neural models outperform n-gram models with more than 1 BLEU point. On the other hand, when memory is not a limitation, traditional n-gram models outperform even the largest neural models by a sizable margin (over 0.5 BLEU in our experiments).

Our work is important because it reviews the most important scaling techniques used in neural language modelling for MT. We show how these methods compare to each other and we combine them to obtain neural models that are fast to both train and test. We conclude by exploring the strengths and weaknesses of these models into greater detail.

\subsubsection*{Acknowledgments}

This work was supported by a Xerox Foundation Award and EPSRC grant number EP/K036580/1.

\bibliographystyle{plainnat}
\bibliography{mybib}

\end{document}